\documentclass[final,5p,times,twocolumn,authoryear]{elsarticle}

\usepackage{amssymb,amsmath}
\usepackage{microtype}
\usepackage{booktabs,tabularx,array,longtable}
\usepackage{flafter}
\usepackage{breakcites}
\usepackage[hidelinks]{hyperref}

\graphicspath{{figures/}}
\journal{Computerized Medical Imaging and Graphics}

\begin{document}

\begin{frontmatter}

\title{Topology-Aware Query Selection for Surgical Instrument Instance Segmentation}

\author[aff1]{Ze Zhang}
\author[aff2]{Yang Zhang}
\affiliation[aff1]{organization={Department of Biomedical Engineering, ShanghaiTech University},city={Shanghai},country={China}}
\affiliation[aff2]{organization={Wuhan United Imaging Surgical Co., Ltd.},city={Wuhan},country={China}}

\begin{abstract}

Accurate foreground masks can still form an incorrect surgical-instrument instance set: duplicate, fragmented, merged, missed, or empty-frame predictions may preserve favorable pixel overlap while violating object identity and count. Final query selection is therefore a relational, variable-cardinality problem rather than a collection of independent candidate decisions. We evaluate topology-aware query selection, which represents the nonempty candidates of a fixed Mask2Former as a complete graph, learns relational candidate and pair representations, predicts set cardinality, and solves an exact structured subset problem. The formal comparison is the complete relational path versus a node-feature-matched path; it evaluates the combined effect of pairwise geometry, message passing, and the additional relational-path capacity, not an isolated component. On the sealed 22-case source test, all three discovery seeds supported instance-set performance improvement with segmentation fidelity and predefined technical-safety preservation: instance F1 increased by 0.0504--0.0612 and positive-frame set-failure rate decreased by 0.0848--0.1060. Direct ROBUST-MIPS transfer reproduced the complete result in all three seeds. Endoscapes supported only one of three seeds and therefore did not establish stable direct transfer. Taken together, the results support a bounded conclusion: the evaluated complete path improved coherent instance-set construction from fixed Mask2Former candidates in specified native-instance contracts, while stable cross-domain transfer and component-specific effects remain unestablished.

\end{abstract}


\end{frontmatter}

\section{Introduction}

Surgical instrument instance segmentation must identify not only where tool foreground lies but also how that foreground is partitioned into individual instruments. This distinction matters whenever downstream analysis follows separate tools, counts them, or assigns tool-specific state. A prediction can preserve nearly the same foreground union while splitting one tool into several masks, merging two tools, duplicating a tool, omitting an instrument, or activating in an empty frame. Surgical datasets and multi-instance methods make object identity an explicit target, but a favorable overlap score alone does not guarantee that the predicted masks form the correct set \citep{kurmann2021mask,ross2021robustmis,angeles2022realtime,alabi2025cholec}.

The final decision over query masks is therefore joint and variable in size. Two candidates may overlap the same instrument and compete for one output slot; one candidate may contain another; adjacent candidates may be fragments of one tool; and the number of valid instances changes across frames. A candidate that appears plausible in isolation can become redundant or inconsistent after the other candidates are considered. These dependencies turn final query selection into structural prediction over a set: the predictor must determine both which candidates belong together in the output and how many candidates the output should contain.

Candidate-wise scoring does not explicitly represent this full relational state. Detector-based models associate masks with region proposals, while DETR-style systems formulate object prediction with learned queries and bipartite matching \citep{he2017mask,carion2020detr}. MaskFormer, Mask2Former, and QueryInst further develop query-based mask prediction and shared instance representations \citep{cheng2021maskformer,cheng2022mask2former,fang2021queryinst}. These advances make strong candidate masks possible. A final unary score still cannot directly encode substantial containment, competition for one reference instance, or a jointly incorrect cardinality. The surgical mask-query problem therefore requires mask topology and variable set size to be represented together.

We address this problem with topology-aware query selection over fixed Mask2Former candidates. The method retains every nonempty candidate and constructs a complete graph whose edges describe overlap, directional containment, relative scale, score separation, and boundary adjacency. Relational message passing updates each candidate in the context of all others, a cardinality head estimates the required output size, and exact structured optimization selects a subset while penalizing learned pair competition. Each module answers a part of the technical problem: pairwise geometry describes candidate conflict, message passing conditions unary evidence on the set, cardinality prediction replaces an implicit threshold count, and exact selection enforces a globally consistent subset.

The formal test compares this complete relational path with a node-feature-matched path that shares the candidates, node descriptors, supervision, folds, training seeds, checkpoint rule, cardinality machinery, and exact selector. The contrast therefore evaluates the complete relational path as a package, including edge geometry, message passing, and its additional capacity. The principal evidence is a one-shot sealed source test, followed by separate direct transfers to ROBUST-MIPS and Endoscapes. The source test and ROBUST-MIPS support the complete method claim across all three discovery seeds, whereas Endoscapes does not establish stable transfer.

The paper makes three contributions. First, it formulates final surgical mask-query selection as relational, variable-cardinality structural prediction. Second, it provides a frozen, case-level matched comparison on a sealed source test and two separately interpreted external domains. Third, it qualifies the method claim under a label-contract-aware, non-compensatory rule: instance-set improvement counts only when segmentation fidelity and predefined technical safety are also preserved within the same evidence setting.

\section{Related work}

\subsection{Surgical instrument instance segmentation}

Surgical instrument segmentation has progressed from foreground and part segmentation toward explicit separation of multiple instrument instances. Multi-instance methods, attention and feature-fusion architectures, and the ROBUST-MIS challenge have highlighted the difficulty of distinguishing nearby or overlapping tools in endoscopic video \citep{kurmann2021mask,ross2021robustmis,angeles2022realtime}. CholecInstanceSeg provides official instrument instance identifiers for laparoscopic cholecystectomy, while Endoscapes and ROBUST-MIPS provide additional native-instance populations with distinct acquisition and annotation conditions \citep{alabi2025cholec,mascagni2025endoscapes,han2026robustmips}. Our focus is narrower than a new backbone or temporal architecture: given fixed mask candidates, we study how the final system should construct a coherent instance set.

\subsection{Query and set prediction}

Mask R-CNN constructs instances from region proposals, whereas DETR treats detection as direct set prediction with object queries and bipartite matching \citep{he2017mask,carion2020detr}. MaskFormer and Mask2Former extend mask classification and masked attention across segmentation tasks, and QueryInst shares query representations across instance attributes \citep{cheng2021maskformer,cheng2022mask2former,fang2021queryinst}. These models already contain interactions within their decoders. Our question begins after the fixed Mask2Former has emitted scored masks: whether the evaluated complete relational path improves the final subset over a model with the same candidate-wise information but no edge geometry or graph messages.

\subsection{Relational reasoning and duplicate control}

Learned suppression can replace hand-crafted NMS by rescoring boxes from their scores and geometry \citep{hosang2017learningnms}. Relation networks use appearance and geometric interactions among detected objects, proposal-ranking methods learn a suppression order, and contextual rescoring conditions each detection on the surrounding detection set \citep{hu2018relation,tan2019rank,pato2020seeing}. These methods establish that candidate interactions can improve box-level ranking or duplicate control.

Our study starts instead from fixed surgical mask queries and asks how to construct a native-instance set with variable cardinality and an exact subset constraint. Relational reasoning, cardinality modeling, and exact optimization are established techniques rather than separate contributions. We integrate them for surgical mask-query topology and evaluate the integrated path under frozen case-level contracts. This design does not claim that its individual components are novel or causally isolated.

\subsection{Evaluation boundary for instance sets}

Instance-segmentation and medical-image-analysis studies have shown that metric choice, aggregation, empty references, and challenge ranking procedures can change the interpretation of model performance \citep{jena2023beyondmap,maierhein2024metrics,reinke2024pitfalls,ostmeier2023use,maierhein2018rankings}. Reporting guidance likewise emphasizes intended use, data partitions, reference standards, and statistical units \citep{tejani2024claim,ross2023beyond}. These considerations are evidence discipline for the present method claim, not a second method contribution. We keep native-instance performance, foreground-union fidelity, empty-reference behavior, and case-level stability separate because none is an interchangeable proxy for the others.

\section{Method}

\begin{figure*}[t]
\centering
\includegraphics[width=\textwidth]{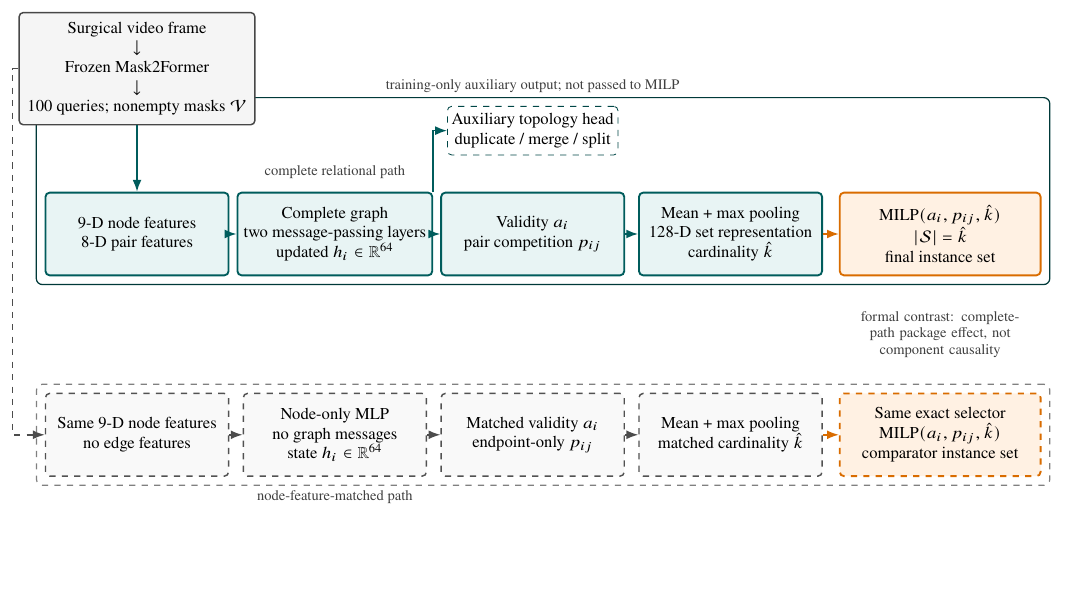}
\caption{Complete-path and matched-comparator data flow. The cardinality head reads the pooled candidate set, and the exact selector receives validity, pair-competition, and cardinality outputs; the auxiliary topology head is used only for training. The formal comparison estimates the complete-path package effect, not the causal effect of any component.}
\label{fig:method}
\end{figure*}

Figure~\ref{fig:method} traces the shared frozen candidates through the two formal paths. The complete path adds pair geometry and graph messages before the matched prediction heads; both paths predict cardinality and use the same exact subset objective.

\subsection{Selecting a coherent set from fixed candidates}

The method begins from fixed candidates so that it changes set construction rather than pixel prediction. For an endoscopic frame $x$, a fixed Mask2Former emits $N=100$ scored mask queries $\mathcal{Q}=\{q_i\}_{i=1}^{N}$ before any external score threshold or mask non-maximum suppression \citep{cheng2022mask2former}. Query $q_i$ contains a tool-class score $s_i$ and binary mask $m_i$. The score retains the no-object class in the softmax denominator. Mask logits are resized bilinearly to $256\times256$, converted to probabilities, and thresholded at 0.5. All queries remain auditable, while only nonempty masks form the eligible set

$\mathcal{V}=\{i:|m_i|>0\}.$

Empty masks create neither graph edges nor selected outputs.

The prediction target is a subset $\mathcal{S}\subseteq\mathcal{V}$ whose masks represent the native instrument instances. Membership is relational: two masks may compete for one reference instance, a candidate may contain another, and adjacent fragments may be inconsistent with one coherent tool. The model therefore predicts both set cardinality and candidate membership without updating the base segmenter.

\subsection{Candidate representation}

Node features describe unary plausibility, candidate geometry, and frame context available before relational inference. Each candidate has nine frozen features: tool score, normalized emitted rank, mask area, area ratio, perimeter, connected-component count, emitted candidate count, empty-query count, and foreground-union area. These features let the matched paths see the same score, scale, shape, and frame-level candidate-density information. The method uses neither image pixels nor temporal, query-embedding, or region-embedding features.

Native-instance annotations define supervision. Hungarian matching at mask IoU $\geq0.5$ supplies node-validity labels. A candidate pair is labeled as competing when both masks overlap the same native reference instance with overlap coefficient $\geq0.5$. Three auxiliary node targets identify candidates associated with duplication, merging, and splitting. Cardinality is represented by classes 0 through 9 and one class for 10 or more instruments. Connected-component proxy labels neither train nor evaluate the method.

\subsection{Complete query graph}

Candidate conflict can involve any pair, so the method does not prune relations before the competition problem is solved. Each frame yields an undirected complete graph $G=(\mathcal{V},\mathcal{E})$ with $n(n-1)/2$ edges for $n=|\mathcal{V}|$. No fixed-neighbor rule or result-dependent edge pruning is used. Every edge has eight features: pairwise intersection over union, two directional containment ratios, overlap coefficient, absolute log area ratio, absolute score margin, minimum boundary distance, and boundary-contact pixels. Together these features distinguish duplicate-like overlap, asymmetric containment, relative scale, confidence separation, and adjacent fragmentation. Means and population standard deviations are estimated only from fit-role cases in each outer fold; a zero standard deviation is replaced by one.

\subsection{Relational inference}

Relational inference revises each unary candidate state using all other eligible candidates and their pair geometry. Two linear layers with ReLU, LayerNorm, and dropout 0.10 map each node to a 64-dimensional state, while two linear layers of widths 32 and 64 encode edge features. Two message-passing layers update node $i$ as

\begin{equation}
\label{eq:message}
\bar r_i^{(\ell)}=\frac{1}{\max(1,|\mathcal{N}(i)|)}\sum_{j\in\mathcal{N}(i)}M_{\ell}\!\left([h_j^{(\ell)},h_i^{(\ell)},e_{ij}]\right).
\end{equation}

\begin{equation}
\label{eq:update}
h_i^{(\ell+1)}=h_i^{(\ell)}+U_{\ell}\!\left([h_i^{(\ell)},\bar r_i^{(\ell)}]\right).
\end{equation}

Equations~\ref{eq:message} and~\ref{eq:update} aggregate degree-normalized pair messages and apply a residual node update. Linear heads then predict node validity $a_i$, the three topology targets, and pair competition $c_{ij}$. Mean and elementwise-maximum pooling summarize the updated candidates into a 128-dimensional set representation for cardinality prediction; an empty eligible set uses a zero pooled representation. Training minimizes the unweighted sum of node-validity, pair-competition, cardinality, and topology losses, using binary cross-entropy with logits except for cardinality cross-entropy.

\subsection{Cardinality prediction}

Independent score thresholds leave output count implicit, although count is part of the instance-set target. The cardinality head therefore predicts

\begin{equation}
\label{eq:cardinality}
\hat{k}=\min\left(\arg\max_{r\in\{0,\ldots,10\}}p(r\mid G),|\mathcal{V}|\right).
\end{equation}

In Equation~\ref{eq:cardinality}, the final class requests 10 candidates before the eligible-set cap. Cardinality is predicted from the relationally pooled candidate representation, not from image pixels, so the output size is conditioned on the candidate configuration rather than inferred by independently thresholding candidates.

\subsection{Exact structured subset selection}

The final selector combines unary validity, learned pair competition, and the predicted cardinality in one constrained decision. With $p_{ij}=\sigma(c_{ij})$, it returns

\begin{equation}
\label{eq:selector}
\mathcal{S}^{*}=\arg\max_{\mathcal{S}\subseteq\mathcal{V},|\mathcal{S}|=\hat{k}}\left[\sum_{i\in\mathcal{S}}a_i-\sum_{i<j;\,i,j\in\mathcal{S}}p_{ij}\right].
\end{equation}

Equation~\ref{eq:selector} fixes the pair penalty at 1.0. Cardinalities one and two use exhaustive enumeration. Larger cardinalities use mixed-integer linear programming with binary node variables $y_i$, pair variables $z_{ij}=y_i y_j$, exact cardinality, standard linearization constraints, and zero relative optimality gap. Candidate identifiers define canonical variable order, and exhaustive exact ties choose the lexicographically earlier subset. An infeasible cardinality, incomplete pair graph, solver failure, or cardinality mismatch creates a flagged unresolved frame; no heuristic subset is substituted.

For $n$ eligible queries, graph construction and relational processing require $O(n^2)$ pair storage and computation at fixed feature widths. The MILP contains $n$ node variables and $n(n-1)/2$ pair variables and has combinatorial worst-case complexity. With $n\leq100$ in this implementation, the pair graph contains at most 4,950 unique edges. Exactness is a design property of both formal paths, not an independently estimated effect in the primary comparison.

\subsection{Node-feature-matched path}

The formal comparator keeps the candidate-wise information and selection machinery fixed. It is a node-only multilayer perceptron that shares candidates, node features, node encoder, prediction heads, pooling, folds, training seeds, optimizer, supervision, checkpoint rule, cardinality prediction, and exact structured selector with the relational path. Its pair head receives concatenated endpoint states but neither edge features nor graph messages. Frozen score-threshold output and mask NMS at 0.50 are secondary controls, not the formal method contrast.

The graph and node-feature-matched paths contain 59,120 and 6,864 trainable parameters, respectively. Consequently, the graph-minus-node-only result evaluates the complete relational path, including pairwise geometry, relational message passing, and additional capacity. It does not establish the separate contribution of edge features, message passing, cardinality prediction, or exact selection.

\subsection{Training and fixed evaluation relation}

Training is case isolated so that candidate relations are learned without using the sealed or external populations. Three fixed Mask2Former discovery models (seeds 17, 23, and 41) provide candidates for 4,246 frames from 18 source-development cases. Six outer folds assign disjoint fit, calibration, and held-out roles: fit cases estimate standardizers and model parameters, calibration cases select checkpoints without updating parameters, and held-out cases are evaluated once after checkpoint selection.

For each discovery seed and outer fold, the graph and node-feature-matched paths are trained independently with graph-training seeds 101, 103, and 107, producing 54 paired fit groups. AdamW uses learning rate $10^{-3}$, weight decay $10^{-4}$, one frame per batch, gradient clipping at 1.0, and 30 epochs. Native-instance F1 is averaged within each calibration case and then equally across cases. The earliest epoch attaining the maximum calibration objective is retained. No architecture, feature, loss weight, optimizer, threshold, or checkpoint search follows held-out inspection.

The sealed source test and the external populations neither train nor calibrate the method. After source development closed, the 54 paired models were applied without modification to the 22-case source test and separately to Endoscapes and ROBUST-MIPS.

\section{Evaluation protocol}

\subsection{Datasets and evidence roles}

The evaluation settings are separated by population, annotation capability, and training/evaluation relation (Table 1). Source development freezes and progresses the method but is not independent confirmation. The sealed source test is the primary one-shot confirmation. Endoscapes and ROBUST-MIPS are distinct direct external transfers and are never combined.

\begin{table*}[t]

\centering

\scriptsize

\caption{Main evaluation settings. The case or video is the inferential unit, and external domains remain separate.}
\label{tab:settings}

\begin{tabularx}{\textwidth}{@{}>{\raggedright\arraybackslash}p{0.20\textwidth}>{\centering\arraybackslash}p{0.12\textwidth}>{\raggedright\arraybackslash}p{0.17\textwidth}>{\raggedright\arraybackslash}X>{\raggedright\arraybackslash}p{0.19\textwidth}@{}}

\toprule

Setting & Frames/\hspace{0pt}cases & Label capability & Training/\hspace{0pt}evaluation relation & Evidence role \\

\midrule

Cholec\hspace{0pt}Instance\hspace{0pt}Seg development & 4,246/18 & Official instance IDs & Out-of-fold method development & Method freezing and progression \\

Cholec\hspace{0pt}Instance\hspace{0pt}Seg test & 10,566/22 & Official instance IDs & Fixed source models applied once & Sealed source confirmation \\

Endoscapes official \texttt{\detokenize{test_seg}} & 74/10 & Native instances & Fixed source models; direct transfer & External target-domain evidence \\

ROBUST-MIPS Testing Stages 1--3 & 4,057/30 & Native instances & Fixed source models; direct transfer & External target-domain evidence \\

\bottomrule

\end{tabularx}

\end{table*}

\subsection{Matched comparison and case-level inference}

The primary estimands are graph minus node-feature-matched native-instance F1 and node-feature-matched minus graph positive-frame set-failure rate. A positive frame is a set failure unless the selected set contains exactly one matched prediction for every reference instance and no false-positive or false-negative instance. Count mean absolute error and duplicate, merge, and split events are supporting outcomes.

Frame outcomes are averaged within each case, and cases contribute equally. Direct-test and external resamples preserve graph/node-only pairs, take the median across six outer-fold models within each graph-training seed, and then take the median across graph-training seeds 101, 103, and 107. Uncertainty uses 10,000 paired case-bootstrap replicates with seed 20260729. The two superiority endpoints form one Holm-adjusted family. Leave-one-case-out analysis tests whether either superiority direction reverses after any one case is removed.

\subsection{Non-compensatory qualification of the method claim}

The complete claim has three non-interchangeable parts. \textbf{Instance-set performance improvement} requires improvement in native-instance F1 and reduction in positive-frame set failure. \textbf{Segmentation fidelity preservation} requires that graph-minus-node-only union Dice satisfies its noninferiority condition. \textbf{Safety preservation} is technical rather than clinical: it requires no increase in false activation on reference-empty frames and zero unresolved-selection flags. Evidence stability is assessed through the predefined seed-replication and influence rules; it is not an extra score that can offset another dimension.

Each discovery seed had to satisfy all predefined conditions. F1 improvement had to be at least 0.03 with a 95\% confidence-interval lower bound above zero, and set-failure reduction had to be at least 0.05 with its lower bound above zero. Both Holm-adjusted one-sided $p$-values had to be below 0.05. The union-Dice confidence-interval lower bound had to exceed $-0.01$, neither superiority direction could reverse under leave-one-case-out analysis, empty-frame false activation could not increase, and unresolved solver flags had to remain zero. Setting-level support requires complete support in at least two of the three fixed discovery seeds. Criteria and thresholds are identical across the direct settings.

The rule is non-compensatory: every applicable dimension must be satisfied within the same seed before that seed counts toward setting-level support. A favorable instance endpoint cannot replace unestablished fidelity or safety preservation, and a successful result in one domain cannot repair a failed result in another. Failure to establish safety preservation denotes an absence of support for preservation; it is not evidence that safety degradation occurred. No completed setting received the latter interpretation.

\subsection{Label-contract boundary}

Native instance identifiers preserve object grouping and support one-to-one matching, instance F1, count, and duplicate/merge/split interpretation. An 8-connected-component proxy derived from semantic foreground preserves only the component structure of the union and need not encode the same object identities. Results are therefore interpreted within their label contract. The proxy analysis uses identical predictions to examine observability; it neither trains nor evaluates the graph method under proxy references.

\section{Results}

\subsection{Primary confirmation on the sealed source test}

The sealed source test provides the strongest confirmation of the complete relational path. All three discovery seeds satisfied the predefined instance-set, fidelity, influence, empty-frame, and solver conditions (Table 2). Native-instance F1 increased by 0.0504 (95\% CI 0.0382--0.0660), 0.0612 (0.0356--0.1038), and 0.0513 (0.0441--0.0770) for seeds 17, 23, and 41. Positive-frame set-failure rate decreased by 0.1060 (0.0825--0.1284), 0.0848 (0.0665--0.0996), and 0.0906 (0.0764--0.1133), respectively.

The instance-set gains were accompanied by segmentation fidelity and technical-safety preservation. Union Dice increased by 0.0450, 0.0551, and 0.0556, with all confidence intervals above the frozen noninferiority margin. Empty-frame false activation changes were -0.0736, -0.0901, and -0.0496; leave-one-case-out minima remained positive for both superiority endpoints; and no unresolved solver flag occurred. Both superiority tests had Holm-adjusted one-sided $p=0.00019998$ for every seed. Together, these observations satisfied the predefined complete criteria on the sealed source test.

\begin{table*}[t]

\centering

\scriptsize

\caption{Frozen case-level comparison of the complete relational path with the node-feature-matched path. Values are displayed to four decimals from the setting-specific decisions and receipts. $\Delta$F1 and $\Delta$Dice are graph minus node-only; set-failure reduction is node-only minus graph; empty $\Delta$ is graph minus node-only false activation. ``Complete'' indicates that every predefined superiority, fidelity, leave-one-case-out, empty-frame, and solver condition was satisfied for that seed.}
\label{tab:main-results}

\setlength{\tabcolsep}{2.7pt}
\begin{tabularx}{\textwidth}{@{}>{\raggedright\arraybackslash}p{0.13\textwidth}>{\centering\arraybackslash}p{0.045\textwidth}>{\centering\arraybackslash}p{0.145\textwidth}>{\centering\arraybackslash}p{0.20\textwidth}>{\centering\arraybackslash}p{0.145\textwidth}>{\centering\arraybackslash}p{0.075\textwidth}>{\centering\arraybackslash}X@{}}

\toprule

Setting & Seed & \shortstack{$\Delta$F1\\95\% CI} & \shortstack{Set-failure reduction\\95\% CI} & \shortstack{$\Delta$Dice\\95\% CI} & Empty $\Delta$ & Complete \\

\midrule

Source test & 17 & \shortstack{.0504\\ {[.0382, .0660]}} & \shortstack{.1060\\ {[.0825, .1284]}} & \shortstack{.0450\\ {[.0310, .0632]}} & -.0736 & Yes \\

Source test & 23 & \shortstack{.0612\\ {[.0356, .1038]}} & \shortstack{.0848\\ {[.0665, .0996]}} & \shortstack{.0551\\ {[.0344, .0982]}} & -.0901 & Yes \\

Source test & 41 & \shortstack{.0513\\ {[.0441, .0770]}} & \shortstack{.0906\\ {[.0764, .1133]}} & \shortstack{.0556\\ {[.0364, .0807]}} & -.0496 & Yes \\

ROBUST-MIPS & 17 & \shortstack{.0346\\ {[.0259, .0484]}} & \shortstack{.1160\\ {[.1001, .1439]}} & \shortstack{.0224\\ {[.0143, .0322]}} & -.0353 & Yes \\

ROBUST-MIPS & 23 & \shortstack{.0714\\ {[.0612, .0859]}} & \shortstack{.1402\\ {[.1190, .1672]}} & \shortstack{.0633\\ {[.0470, .0772]}} & -.1106 & Yes \\

ROBUST-MIPS & 41 & \shortstack{.0459\\ {[.0383, .0552]}} & \shortstack{.1357\\ {[.1159, .1630]}} & \shortstack{.0378\\ {[.0273, .0478]}} & -.0459 & Yes \\

Endoscapes & 17 & \shortstack{.0211\\ {[-.0131, .0633]}} & \shortstack{.1284\\ {[.0332, .2313]}} & \shortstack{.0093\\ {[-.0173, .0458]}} & +.0029 & No \\

Endoscapes & 23 & \shortstack{.0586\\ {[.0280, .0937]}} & \shortstack{.0984\\ {[.0561, .1519]}} & \shortstack{.0545\\ {[.0065, .1075]}} & .0000 & Yes \\

Endoscapes & 41 & \shortstack{.0234\\ {[-.0197, .0728]}} & \shortstack{.0913\\ {[.0211, .1917]}} & \shortstack{.0320\\ {[-.0043, .0877]}} & .0000 & No \\

\bottomrule

\end{tabularx}

\end{table*}

\begin{figure*}[t]
\centering
\includegraphics[width=\textwidth]{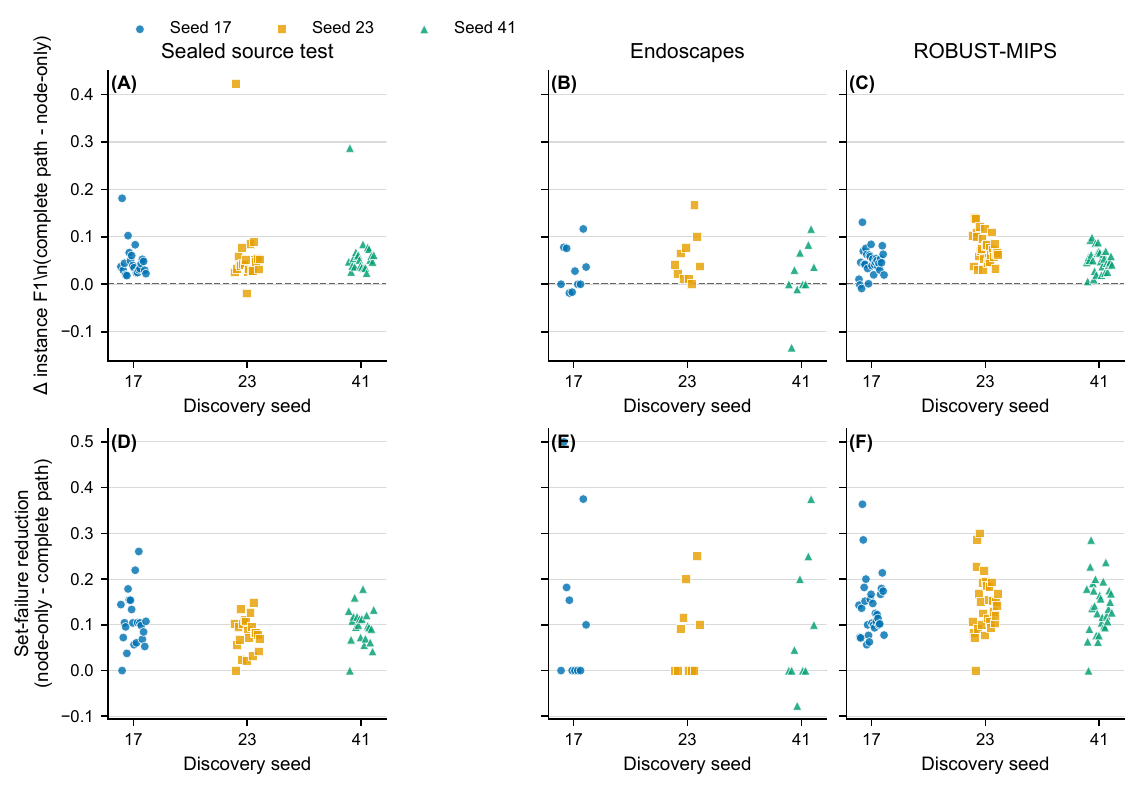}
\caption{Case-level differences in the two superiority endpoints. Panels A--C show complete-path minus node-feature-matched instance F1; panels D--F show node-feature-matched minus complete-path set-failure rate. Points are cases or videos within discovery seed after the visualization-only nested median. The display is descriptive; Table~\ref{tab:main-results} and frozen bootstrap receipts are confirmatory, and domains are not pooled.}
\label{fig:case-distributions}
\end{figure*}

Figure~\ref{fig:case-distributions} shows the case-level directions underlying the frozen setting summaries. The broadest and least consistent F1 spread occurs in Endoscapes; the domains remain descriptive panels rather than a pooled analysis.

Source development preceded this confirmation and served a different role. It supported method progression in two of three discovery seeds. Seeds 17 and 23 satisfied the complete development criteria. Seed 41 had an F1 difference of 0.0297 (95\% CI 0.0087--0.0507), below the predefined 0.03 point criterion, and an empty-frame false-activation increase of 0.0267. The later sealed result confirms the method under the source-test relation without rewriting that earlier development uncertainty.

\subsection{Direct external transfer}

\subsubsection{ROBUST-MIPS}

Direct ROBUST-MIPS transfer reproduced the complete result in all three discovery seeds. Native-instance F1 increased by 0.0346--0.0714. Positive-frame set-failure rate decreased by 0.1160--0.1402. Union-Dice changes were 0.0224--0.0633. Both superiority endpoints had Holm-adjusted one-sided $p=0.00019998$ for each seed, leave-one-case-out directions remained positive, empty-frame false activation did not increase, and solver flags were zero. These results satisfied the complete predefined criteria within the ROBUST-MIPS contract.

The replication remains dataset specific. ROBUST-MIPS did not train, calibrate, or select the relational path, but the result does not imply that every external surgical domain will reproduce the effect or that the method is prospectively confirmed across all external data.

\subsubsection{Endoscapes}

Endoscapes was seed-unstable. Seed 23 satisfied every predefined component, with F1 improvement 0.0586 (95\% CI 0.0280--0.0937), set-failure reduction 0.0984 (0.0561--0.1519), union-Dice change 0.0545 (0.0065--0.1075), no increase in empty-frame false activation, and zero solver flags. Seeds 17 and 41 did not meet the F1 point and confidence requirements. Seed 17 additionally had a union-Dice confidence lower bound of -0.0173, below the -0.01 margin, and an empty-frame false-activation increase of 0.0029. Only one of three seeds therefore supported the complete claim.

Set-failure rate fell in all three Endoscapes seeds, but the full criterion held in only one. ROBUST-MIPS cannot compensate for that result because the populations and evidence contracts are distinct. The two settings therefore remain separate: ROBUST-MIPS supports successful direct transfer, whereas Endoscapes remains seed-unstable rather than supporting a pooled external conclusion.

\subsection{What is preserved when the instance set improves}

The sealed source test and ROBUST-MIPS paired instance-set improvement with preservation of the foreground union and the predefined technical-safety variables. In both settings, every seed that counted toward the method claim also preserved union Dice under the frozen margin, avoided increased empty-reference activation, and produced zero unresolved states. Segmentation fidelity preservation and safety preservation are therefore part of the positive method result rather than separate scorecards.

That conjunction matters in Endoscapes. Seed 17 reduced set failure, but did not establish the required F1 improvement, fidelity preservation, or empty-frame preservation. For this seed, the supported interpretation is narrower: safety preservation was not established. The analysis did not test or support the stronger conclusion that safety degradation occurred.

\subsection{Dependence on the label contract}

Identical predictions produced different instance conclusions when native object identities were replaced with 8-connected components of the same foreground union (Supplement Section 4, label-contract audit). In CholecInstanceSeg development, union Dice was exactly invariant by construction. All three Mask R-CNN seeds nevertheless had lower instance F1 and higher count error under the proxy contract, with case-level intervals excluding zero. The SegFormer-plus-components model was nearly unchanged in aggregate because its predictions already used the same component ontology.

The frozen post-hoc Endoscapes replication showed the same observability boundary. Union Dice remained invariant, while instance F1 decreased and count error increased for all four candidate bundles. Only 6 of 74 frames changed topology under the native-to-proxy conversion; the 68 unchanged-count frames had exactly zero contract deltas. The effect is therefore localized to the reference grouping change rather than to different predictions or image populations.

These analyses explain why foreground-union similarity does not establish native-instance identity and why cross-dataset evidence must retain its label contract. They do not evaluate the topology-aware selector against proxy references. The full native-versus-proxy numeric ledger, label-contract figure, and setting-level claim matrix are in Supplement Sections 4 and 5.

\begin{table*}[t]
\centering
\small
\caption{Frozen component-level runtime profile on 300 source-development frames using an NVIDIA RTX 4070 SUPER. Timings are per frame and do not include Mask2Former inference or data loading.}
\label{tab:runtime}
\begin{tabularx}{\textwidth}{@{}>{\raggedright\arraybackslash}Xrr>{\raggedright\arraybackslash}p{0.36\textwidth}@{}}
\toprule
Component & Median (ms) & 95th percentile (ms) & Measurement scope \\
\midrule
Complete-graph construction & 390.9 & 435.8 & Node/pair feature assembly for eligible candidates \\
Complete-path forward pass & 0.98 & 2.30 & Relational network and prediction heads \\
Node-only forward pass & 0.83 & 0.87 & Matched node-only network and heads \\
Exact subset selector & 23.0 & 277.9 & Exhaustive or zero-gap MILP selection \\
\bottomrule
\end{tabularx}
\end{table*}

The profiled frames contained a median of 96 eligible nodes and 4,560 unique pairs; the respective 95th percentiles were 99 and 4,851. Peak recorded memory was 26.4 MB for the complete path and 15.1 MB for the node-only path. These isolated measurements characterize implementation cost under one hardware and candidate regime. They are not end-to-end latency measurements and do not establish real-time operation.

\subsection{Supplemental boundary evidence}

The supplemental leave-one-domain-out study tested broader held-out robustness, but did not support a general multi-domain claim. Under its distinct post-result training relation and graph seed 101, complete support occurred in one of three source-held-out seeds, zero of three Endoscapes-held-out seeds, one of three ROBUST-MIPS Stage-2 seeds, and two of three Stage-3 seeds. These results bound generality but do not replace the one-shot source test or the direct external evaluations.

The A2 supervision analysis also limits component attribution. It ended after prospectively amended Stage A at graph seed 101. Removing pair-competition supervision gave $\Delta$F1 0.0057 (95\% CI -0.0102--0.0182), set-failure reduction 0.0121 (-0.0071--0.0374), and union-Dice change -0.0020 (-0.0141--0.0099), producing an unstable Stage-A interpretation. Removing cardinality supervision gave $\Delta$F1 -0.0118 (-0.0181---0.0030), set-failure reduction -0.0068 (-0.0170--0.0039), and union-Dice change -0.0022 (-0.0125--0.0094), providing no Stage-A support. Stage B was permanently cancelled under the prospective budget amendment and was not evaluated; these data do not isolate the contribution of either component.

Finally, the LODO source-held-out empty-frame follow-up is descriptive post-hoc evidence. Seed 23 showed a distributed positive-excess pattern and seed 41 a concentrated pattern, with low frame-level overlap between their graph-only activations. The frozen fields do not causally identify cardinality, domain shift, or another mechanism, and no confirmatory probability value was available. The audit therefore leaves the affected state as safety preservation not established, not supported safety degradation.

\section{Discussion}

The evaluated complete graph path improved the assembly of fixed surgical mask candidates into an instance set in the supported settings. The design jointly represents candidate geometry, relational context, output cardinality, and structured subset selection. The sealed source test provides the principal confirmation. The complete-path result was reproduced on ROBUST-MIPS under the direct-transfer protocol while preserving foreground fidelity and predefined technical safety.

The matched comparator determines the resolution of this conclusion. Both paths receive the same candidates and node features, use the same supervision and training roles, predict cardinality, and invoke the same exact selector. The complete path also contains edge geometry, graph messages, and substantially greater model capacity. The primary comparison therefore identifies a complete-path effect that includes relational representation, cardinality behavior, selector operation, and capacity differences. It does not identify the separate contribution of the exact selector, the cardinality head, individual edge features, or message passing. The Stage-A-only A2 results do not change this boundary.

External behavior varied by domain. ROBUST-MIPS reproduced the complete conclusion in all seeds, whereas Endoscapes reproduced it in only one. Differences in sample size, candidate behavior, visual domain, and annotation practice may all be relevant, but the completed evidence does not identify a causal explanation. Label capability provides one clear interpretive constraint: a foreground union can be identical while the reference grouping and observable instance identity change. Population and label contracts must therefore remain visible when the method is transported.

The non-compensatory evaluation keeps these outcomes together. Instance F1 and set failure test whether the instance set improves; union Dice tests whether foreground segmentation is preserved; empty-reference activation and unresolved states test predefined technical-safety behavior. A claim is supported only when these properties align within the same seed and setting. This rule explains why favorable Endoscapes set-failure directions do not establish stable transfer and why an unestablished preservation condition is not automatically evidence of degradation.

Exploratory mechanism controls were consistent with a substantial cardinality contribution but did not identify a stable fixed-cardinality relation-specific residual. Capacity-matched, shuffled, high-resolution, calibration, and presence controls also failed to establish a stable joint performance-and-safety condition. These pressure tests do not alter the formal complete-path result; their designs and values are reported in the Supplement.

Several limitations bound the result. The method operates on candidates from one fixed Mask2Former family, so detector-family and architecture generality are untested. The complete graph is quadratic in eligible-candidate count, exact optimization has combinatorial worst-case complexity, and the component timings in Table~\ref{tab:runtime} do not establish real-time deployment. Only two direct external domains were evaluated, and Endoscapes contains 10 video units with correspondingly wide uncertainty. The supplemental multi-domain study did not support a general held-out robustness claim. Technical safety here covers empty-reference activation and unresolved selection states, not clinical safety, clinical benefit, workflow utility, or patient outcome. Component effects remain unresolved because A2 stopped after Stage A, and the empty-frame follow-up remains post hoc and descriptive.

\section{Conclusion}

Good foreground masks can still form the wrong surgical-instrument instance set because candidate validity depends on relations and variable cardinality. Topology-aware query selection addresses this problem with a complete candidate graph, relational inference, cardinality prediction, and exact structured subset selection. Relative to a node-feature-matched path, the complete relational path was supported on the sealed source test and directly replicated on ROBUST-MIPS while preserving segmentation fidelity and predefined technical safety. Endoscapes did not establish stable direct transfer, and supplemental evidence did not support general multi-domain held-out robustness or component-specific attribution. The evidence supports a bounded claim: the evaluated complete topology-aware path improved coherent instance-set construction from fixed Mask2Former candidates under the evaluated native-instance contracts, but its broader transfer and any component-specific effect remain unestablished.

\section*{Data and code availability}

The datasets remain governed by their original providers and cited source publications. This manuscript does not alter those access conditions. Analysis code, fixed decision files, and audit receipts used to prepare this submission are available at \url{https://github.com/yqjyzzz/structured-instance-set-selection}. The underlying datasets are not redistributed and must be obtained from their respective official providers.

\section*{Ethics statement}

No prospective human-subject recruitment or intervention was performed for this methodological analysis. Dataset-specific ethics and consent information is provided by the original dataset publications.

\section*{Declaration of generative AI and AI-assisted technologies in the writing process}

During manuscript preparation, the authors used OpenAI Codex to assist with language drafting, document organization, consistency checking, and typesetting. The authors reviewed and edited the resulting content and take responsibility for the manuscript.

\section*{Declaration of competing interest}

Yang Zhang is employed by Wuhan United Imaging Surgical Co., Ltd. The company had no role in the study design, data analysis, interpretation of results, or decision to submit the manuscript. The authors declare no other competing interests.

\section*{Funding}

This research received no specific grant from any funding agency in the public, commercial, or not-for-profit sectors.

\bibliographystyle{elsarticle-harv}

\bibliography{references}

@article{kurmann2021mask,
  author = {Kurmann, Thomas and M{\'a}rquez-Neila, Pablo and Allan, Max and Wolf, Sebastian and Sznitman, Raphael},
  title = {Mask then classify: multi-instance segmentation for surgical instruments},
  journal = {International Journal of Computer Assisted Radiology and Surgery},
  volume = {16},
  number = {7},
  pages = {1227--1236},
  year = {2021},
  doi = {10.1007/s11548-021-02404-2}
}

@article{ross2021robustmis,
  author = {Ro{\ss}, Tobias and Reinke, Annika and Full, Peter M. and others},
  title = {Comparative validation of multi-instance instrument segmentation in endoscopy: Results of the {ROBUST-MIS} 2019 challenge},
  journal = {Medical Image Analysis},
  volume = {70},
  pages = {101920},
  year = {2021},
  doi = {10.1016/j.media.2020.101920}
}

@article{angeles2022realtime,
  author = {{\'A}ngeles Cer{\'o}n, Juan Carlos and Ochoa Ruiz, Gilberto and Chang, Leonardo and Ali, Sharib},
  title = {Real-time instance segmentation of surgical instruments using attention and multi-scale feature fusion},
  journal = {Medical Image Analysis},
  volume = {81},
  pages = {102569},
  year = {2022},
  doi = {10.1016/j.media.2022.102569}
}

@article{alabi2025cholec,
  author = {Alabi, Oluwatosin and Toe, Ko Ko Zayar and Zhou, Zijian and Budd, Charlie and Raison, Nicholas and Shi, Miaojing and Vercauteren, Tom},
  title = {{CholecInstanceSeg}: A Tool Instance Segmentation Dataset for Laparoscopic Surgery},
  journal = {Scientific Data},
  volume = {12},
  number = {1},
  pages = {825},
  year = {2025},
  doi = {10.1038/s41597-025-05163-w}
}

@inproceedings{carion2020detr,
  author = {Carion, Nicolas and Massa, Francisco and Synnaeve, Gabriel and Usunier, Nicolas and Kirillov, Alexander and Zagoruyko, Sergey},
  title = {End-to-End Object Detection with Transformers},
  booktitle = {European Conference on Computer Vision},
  pages = {213--229},
  year = {2020},
  doi = {10.1007/978-3-030-58452-8_13}
}

@inproceedings{cheng2021maskformer,
  author = {Cheng, Bowen and Schwing, Alexander G. and Kirillov, Alexander},
  title = {Per-Pixel Classification is Not All You Need for Semantic Segmentation},
  booktitle = {Advances in Neural Information Processing Systems},
  volume = {34},
  pages = {17864--17875},
  year = {2021},
  url = {https://proceedings.neurips.cc/paper/2021/hash/950a4152c2b4aa3ad78bdd6b366cc179-Abstract.html}
}

@inproceedings{cheng2022mask2former,
  author = {Cheng, Bowen and Misra, Ishan and Schwing, Alexander G. and Kirillov, Alexander and Girdhar, Rohit},
  title = {Masked-Attention Mask Transformer for Universal Image Segmentation},
  booktitle = {Proceedings of the IEEE/CVF Conference on Computer Vision and Pattern Recognition},
  pages = {1290--1299},
  year = {2022},
  doi = {10.1109/CVPR52688.2022.00135},
  url = {https://openaccess.thecvf.com/content/CVPR2022/html/Cheng_Masked-Attention_Mask_Transformer_for_Universal_Image_Segmentation_CVPR_2022_paper.html}
}

@inproceedings{fang2021queryinst,
  author = {Fang, Yuxin and Yang, Shusheng and Wang, Xinggang and Li, Yu and Fang, Chen and Shan, Ying and Feng, Bin and Liu, Wenyu},
  title = {Instances As Queries},
  booktitle = {Proceedings of the IEEE/CVF International Conference on Computer Vision},
  pages = {6910--6919},
  year = {2021},
  doi = {10.1109/ICCV48922.2021.00683},
  url = {https://openaccess.thecvf.com/content/ICCV2021/html/Fang_Instances_As_Queries_ICCV_2021_paper.html}
}

@inproceedings{jena2023beyondmap,
  author = {Jena, Rohit and Zhornyak, Lukas and Doiphode, Nehal and Chaudhari, Pratik and Buch, Vivek and Gee, James and Shi, Jianbo},
  title = {Beyond {mAP}: Towards Better Evaluation of Instance Segmentation},
  booktitle = {Proceedings of the IEEE/CVF Conference on Computer Vision and Pattern Recognition},
  pages = {11309--11318},
  year = {2023},
  doi = {10.1109/CVPR52729.2023.01088},
  url = {https://openaccess.thecvf.com/content/CVPR2023/html/Jena_Beyond_mAP_Towards_Better_Evaluation_of_Instance_Segmentation_CVPR_2023_paper.html}
}

@article{maierhein2024metrics,
  author = {Maier-Hein, Lena and Reinke, Annika and Godau, Patrick and others},
  title = {Metrics Reloaded: recommendations for image analysis validation},
  journal = {Nature Methods},
  volume = {21},
  number = {2},
  pages = {195--212},
  year = {2024},
  doi = {10.1038/s41592-023-02151-z}
}

@article{reinke2024pitfalls,
  author = {Reinke, Annika and Tizabi, Minu D. and Baumgartner, Michael and others},
  title = {Understanding metric-related pitfalls in image analysis validation},
  journal = {Nature Methods},
  volume = {21},
  number = {2},
  pages = {182--194},
  year = {2024},
  doi = {10.1038/s41592-023-02150-0}
}

@article{ostmeier2023use,
  author = {Ostmeier, Sophie and Axelrod, Brian and Isensee, Fabian and Bertels, Jeroen and Mlynash, Michael and Christensen, Soren and Lansberg, Maarten G. and Albers, Gregory W. and Sheth, Rajen and Verhaaren, Benjamin F. J. and Mahammedi, Abdelkader and Li, Li-Jia and Zaharchuk, Greg and Heit, Jeremy J.},
  title = {{USE-Evaluator}: Performance metrics for medical image segmentation models supervised by uncertain, small or empty reference annotations in neuroimaging},
  journal = {Medical Image Analysis},
  volume = {90},
  pages = {102927},
  year = {2023},
  doi = {10.1016/j.media.2023.102927}
}

@article{mascagni2025endoscapes,
  author = {Mascagni, Pietro and Alapatt, Deepak and Murali, Aditya and Vardazaryan, Armine and Garcia, Alain and Okamoto, Nariaki and Costamagna, Guido and Mutter, Didier and Marescaux, Jacques and Dallemagne, Bernard and Padoy, Nicolas},
  title = {Endoscapes, a critical view of safety and surgical scene segmentation dataset for laparoscopic cholecystectomy},
  journal = {Scientific Data},
  volume = {12},
  number = {1},
  pages = {331},
  year = {2025},
  doi = {10.1038/s41597-025-04642-4}
}

@article{han2026robustmips,
  author = {Han, Zhe and Budd, Charlie and Zhang, Gongyu and Tian, Huanyu and Bergeles, Christos and Vercauteren, Tom},
  title = {{ROBUST-MIPS}: A Combined Skeletal Pose and Instance Segmentation Dataset for Laparoscopic Surgical Instruments},
  journal = {Scientific Data},
  volume = {13},
  number = {1},
  pages = {684},
  year = {2026},
  doi = {10.1038/s41597-026-06938-5}
}

@inproceedings{he2017mask,
  author = {He, Kaiming and Gkioxari, Georgia and Doll{\'a}r, Piotr and Girshick, Ross},
  title = {Mask {R-CNN}},
  booktitle = {Proceedings of the IEEE International Conference on Computer Vision},
  pages = {2961--2969},
  year = {2017},
  doi = {10.1109/ICCV.2017.322}
}

@article{ross2023beyond,
  author = {Ro{\ss}, Tobias and Bruno, Pierangela and Reinke, Annika and Wiesenfarth, Manuel and Koeppel, Lisa and Full, Peter M. and Pekdemir, B{\"u}nyamin and Godau, Patrick and Trofimova, Darya and Isensee, Fabian and Adler, Tim J. and Tran, Thuy N. and Moccia, Sara and Calimeri, Francesco and M{\"u}ller-Stich, Beat P. and Kopp-Schneider, Annette and Maier-Hein, Lena},
  title = {Beyond rankings: Learning (more) from algorithm validation},
  journal = {Medical Image Analysis},
  volume = {86},
  pages = {102765},
  year = {2023},
  doi = {10.1016/j.media.2023.102765}
}

@article{maierhein2018rankings,
  author = {Maier-Hein, Lena and Eisenmann, Matthias and Reinke, Annika and others},
  title = {Why rankings of biomedical image analysis competitions should be interpreted with care},
  journal = {Nature Communications},
  volume = {9},
  number = {1},
  pages = {5217},
  year = {2018},
  doi = {10.1038/s41467-018-07619-7}
}

@article{tejani2024claim,
  author = {Tejani, Ali S. and Klontzas, Michail E. and Gatti, Anthony A. and Mongan, John T. and Moy, Linda and Park, Seong Ho and Kahn, Charles E. Jr. and {CLAIM 2024 Update Panel}},
  title = {Checklist for Artificial Intelligence in Medical Imaging ({CLAIM}): 2024 Update},
  journal = {Radiology: Artificial Intelligence},
  volume = {6},
  number = {4},
  pages = {e240300},
  year = {2024},
  doi = {10.1148/ryai.240300}
}

@inproceedings{hosang2017learningnms,
  author = {Hosang, Jan and Benenson, Rodrigo and Schiele, Bernt},
  title = {Learning Non-Maximum Suppression},
  booktitle = {Proceedings of the IEEE Conference on Computer Vision and Pattern Recognition},
  pages = {4507--4515},
  year = {2017},
  doi = {10.1109/CVPR.2017.685},
  url = {https://openaccess.thecvf.com/content_cvpr_2017/html/Hosang_Learning_Non-Maximum_Suppression_CVPR_2017_paper.html}
}

@inproceedings{hu2018relation,
  author = {Hu, Han and Gu, Jiayuan and Zhang, Zheng and Dai, Jifeng and Wei, Yichen},
  title = {Relation Networks for Object Detection},
  booktitle = {Proceedings of the IEEE Conference on Computer Vision and Pattern Recognition},
  pages = {3588--3597},
  year = {2018},
  doi = {10.1109/CVPR.2018.00378},
  url = {https://openaccess.thecvf.com/content_cvpr_2018/html/Hu_Relation_Networks_for_CVPR_2018_paper.html}
}

@inproceedings{tan2019rank,
  author = {Tan, Zhiyu and Nie, Xuecheng and Qian, Qi and Li, Nan and Li, Hao},
  title = {Learning to Rank Proposals for Object Detection},
  booktitle = {Proceedings of the IEEE/CVF International Conference on Computer Vision},
  pages = {8273--8281},
  year = {2019},
  doi = {10.1109/ICCV.2019.00836},
  url = {https://openaccess.thecvf.com/content_ICCV_2019/html/Tan_Learning_to_Rank_Proposals_for_Object_Detection_ICCV_2019_paper.html}
}

@inproceedings{pato2020seeing,
  author = {Pato, Louren{\c c}o V. and Negrinho, Renato M. P. and Aguiar, Pedro M. Q.},
  title = {Seeing without Looking: Contextual Rescoring of Object Detections for {AP} Maximization},
  booktitle = {Proceedings of the IEEE/CVF Conference on Computer Vision and Pattern Recognition},
  pages = {14610--14618},
  year = {2020},
  doi = {10.1109/CVPR42600.2020.01462},
  url = {https://openaccess.thecvf.com/content_CVPR_2020/html/Pato_Seeing_without_Looking_Contextual_Rescoring_of_Object_Detections_for_AP_CVPR_2020_paper.html}
}

\clearpage
\onecolumn
\appendix
\renewcommand{\thetable}{S\arabic{table}}
\renewcommand{\thefigure}{S\arabic{figure}}
\renewcommand{\theHtable}{S\arabic{table}}
\renewcommand{\theHfigure}{S\arabic{figure}}
\setcounter{table}{0}
\setcounter{figure}{0}
\newcolumntype{Y}{>{\raggedright\arraybackslash}X}
\section*{Supplementary Material}
\addcontentsline{toc}{section}{Supplementary Material}

\section{Scope and evidence roles}
This supplement provides traceability for the frozen formal analyses and records exploratory mechanism diagnostics. It does not change any model, threshold, seed, sample definition, estimand, label contract, or machine decision. Source development, direct source test, Endoscapes, and ROBUST-MIPS remain separate settings.

Formal evidence is confirmatory for the bounded system claim. Mechanism reports are exploratory pressure tests. They may constrain interpretation, but they cannot upgrade a formal result or prove that one component caused the complete-path difference.

\section{Frozen evaluation protocol}
The unit of analysis is the case or video. Frame outcomes are averaged within case, cases contribute equally, paired graph/node-only outputs are preserved, and the direct-test/external estimand takes the median across six source folds within each graph-training seed and then the median across graph-training seeds 101, 103, and 107. Uncertainty uses 10,000 paired case-bootstrap replicates with seed 20260729. Discovery seeds are 17, 23, and 41. There are 54 atomic pairs per direct setting.

For a seed to count, instance F1 improvement is at least 0.03 with a 95\% interval lower bound above zero; positive-frame set-failure reduction is at least 0.05 with a lower bound above zero; both Holm-adjusted one-sided superiority p-values are below 0.05; the union-Dice lower bound is above -0.01; neither superiority endpoint reverses under leave-one-case-out analysis; empty-frame false activation does not increase; and unresolved solver flags are zero. A setting is supported when the complete result occurs in at least two of three discovery seeds.

Failure to establish safety preservation means that preservation was not supported for the relevant analysis. It is not evidence that safety degradation occurred; the formal analyses did not reach that stronger conclusion.

\section{Formal evidence ledger}
\subsection{Setting scope}
\scriptsize
\begin{longtable}{@{}p{0.19\textwidth}rrrrp{0.20\textwidth}p{0.24\textwidth}@{}}
\caption{Main evaluation settings and evidence roles.}\label{tab:s1}\\
\toprule
Setting & Frames & Cases & Pairs & Label contract & Relation & Decision\\
\midrule
\endhead
CholecInstanceSeg development & 4,246 & 18 & -- & Native instance IDs & Out-of-fold development & Method progressed in 2/3 seeds\\
CholecInstanceSeg source test & 10,566 & 22 & 54 & Native instance IDs & Fixed source models applied once & Complete support in 3/3 seeds\\
Endoscapes \texttt{test\_seg} & 74 & 10 & 54 & Native instances & Fixed source models; direct transfer & Seed-unstable support (1/3)\\
ROBUST-MIPS Testing Stages 1--3 & 4,057 & 30 & 54 & Native instances & Fixed source models; direct transfer & Complete support in 3/3 seeds\\
\bottomrule
\end{longtable}
\normalsize

\subsection{Source development}
\refstepcounter{table}\noindent\textbf{Table \thetable.} Source development by seed.
\begin{tabular}{@{}rrrrr l@{}}
\toprule
Seed & $\Delta$F1 & Set-failure reduction & $\Delta$Dice & Empty $\Delta$ & Complete\\
\midrule
17 & 0.0610 & 0.1281 & 0.0645 & -0.1251 & Yes\\
23 & 0.0607 & 0.0956 & 0.0599 & -0.1852 & Yes\\
41 & 0.0297 & 0.0666 & 0.0263 & +0.0267 & No\\
\bottomrule
\end{tabular}

The source-development decision is supported by 2/3 seeds. Seed 41 is not pooled with the two supporting seeds: its F1 difference is below the 0.03 point criterion and empty false activation increased.

\subsection{Direct source test}
\scriptsize
\begin{longtable}{@{}r p{0.16\textwidth} p{0.18\textwidth} p{0.15\textwidth} r r p{0.14\textwidth}@{}}
\caption{Direct source-test decisions by discovery seed.}\label{tab:s3}\\
\toprule
Seed & $\Delta$F1 [95\% CI] & Set-failure reduction [95\% CI] & $\Delta$Dice [95\% CI] & Empty $\Delta$ & Holm p & LOCO F1 / set\\
\midrule
\endhead
17 & 0.0504 [0.0382, 0.0660] & 0.1060 [0.0825, 0.1284] & 0.0450 [0.0310, 0.0632] & -0.0736 & 0.00019998 & 0.04696 / 0.10070\\
23 & 0.0612 [0.0356, 0.1038] & 0.0848 [0.0665, 0.0996] & 0.0551 [0.0344, 0.0982] & -0.0901 & 0.00019998 & 0.04406 / 0.07680\\
41 & 0.0513 [0.0441, 0.0770] & 0.0906 [0.0764, 0.1133] & 0.0556 [0.0364, 0.0807] & -0.0496 & 0.00019998 & 0.04909 / 0.08847\\
\bottomrule
\end{longtable}
\normalsize
The direct source-test estimand takes the median of the six source-fold equal-case means within each graph-training seed and then the median across graph-training seeds. All three discovery seeds supported the complete source-test conclusion.

\subsection{Direct external settings}
\begin{longtable}{@{}l r r r r r l@{}}
\caption{Direct external-setting formal decisions.}\label{tab:s4}\\
\toprule
Setting & Seed & $\Delta$F1 & Set-failure reduction & $\Delta$Dice & Empty $\Delta$ & Complete\\
\midrule
\endhead
Endoscapes & 17 & 0.0211 & 0.1284 & 0.0093 & +0.0029 & No\\
Endoscapes & 23 & 0.0586 & 0.0984 & 0.0545 & 0.0000 & Yes\\
Endoscapes & 41 & 0.0234 & 0.0913 & 0.0320 & 0.0000 & No\\
ROBUST-MIPS & 17 & 0.0346 & 0.1160 & 0.0224 & -0.0353 & Yes\\
ROBUST-MIPS & 23 & 0.0714 & 0.1402 & 0.0633 & -0.1106 & Yes\\
ROBUST-MIPS & 41 & 0.0459 & 0.1357 & 0.0378 & -0.0459 & Yes\\
\bottomrule
\end{longtable}
Endoscapes support is 1/3, so stable direct transfer was not established. ROBUST-MIPS support is 3/3. The settings are not pooled.

\section{Label-contract audit}
The native-versus-8-connected-component analysis applies identical predictions to two reference constructions. Union Dice is invariant by construction, while instance grouping and count can change. These contrasts are interpretive and do not evaluate the graph method under proxy labels.

\begin{figure}[t]
\centering
\includegraphics[width=\textwidth]{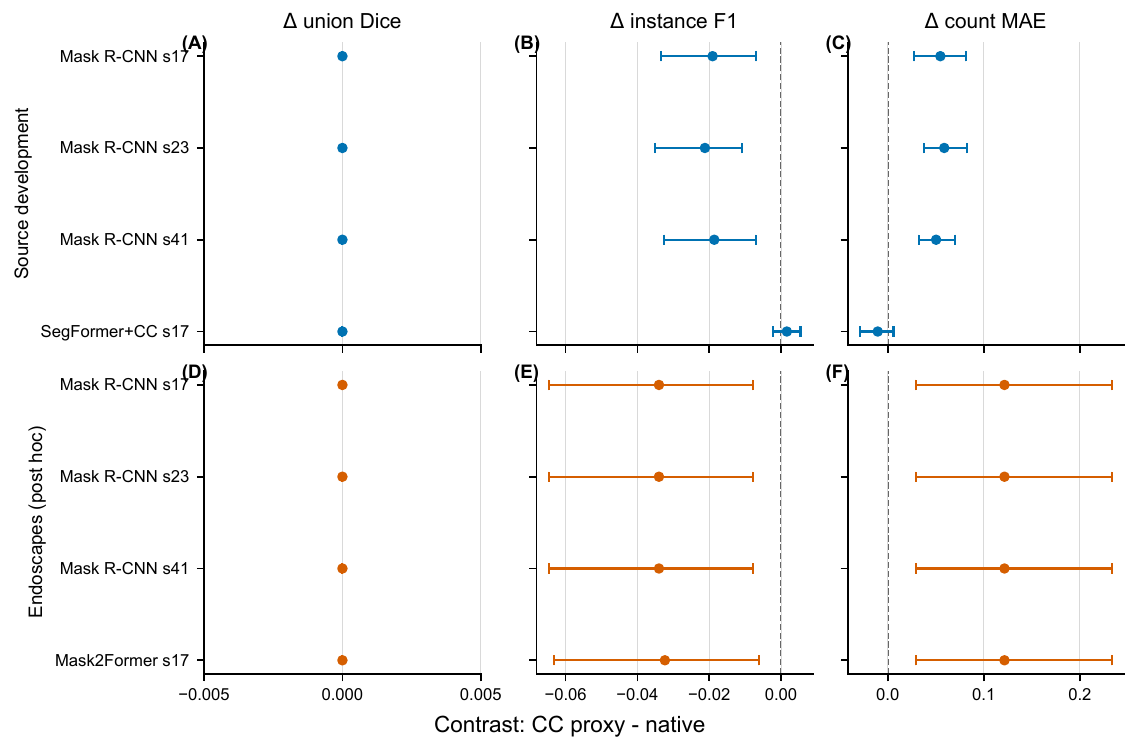}
\caption{Label-contract dependence under identical predictions. Union-Dice differences remain zero by construction, whereas native-to-proxy changes in instance F1 and count error appear where reference grouping changes. The figure is an observability audit, not an evaluation of the selector under proxy labels.}
\label{fig:s1-label-contract}
\end{figure}

\scriptsize
\begin{longtable}{@{}p{0.17\textwidth}p{0.15\textwidth} r p{0.17\textwidth} p{0.19\textwidth}@{}}
\caption{Native-versus-proxy label-contract audit.}\label{tab:s5}\\
\toprule
Population & Model & $\Delta$ union Dice & $\Delta$ instance F1 [95\% CI] & $\Delta$ count MAE [95\% CI]\\
\midrule
\endhead
Cholec development & Mask R-CNN s17 & 0.0000 & -0.0191 [-0.0334, -0.0069] & +0.0548 [+0.0277, +0.0813]\\
Cholec development & Mask R-CNN s23 & 0.0000 & -0.0212 [-0.0350, -0.0110] & +0.0588 [+0.0376, +0.0823]\\
Cholec development & Mask R-CNN s41 & 0.0000 & -0.0186 [-0.0325, -0.0070] & +0.0503 [+0.0321, +0.0703]\\
Cholec development & SegFormer s17 & 0.0000 & +0.0016 [-0.0022, +0.0054] & -0.0104 [-0.0285, +0.0060]\\
Endoscapes & Mask R-CNN s17/23/41 & 0.0000 & -0.0340 [-0.0646, -0.0079] & +0.1215 [+0.0291, +0.2333]\\
Endoscapes & Mask2Former s17 & 0.0000 & -0.0323 [-0.0632, -0.0062] & +0.1215 [+0.0291, +0.2333]\\
\bottomrule
\end{longtable}
\normalsize

\section{Setting-level claim matrix}
\small
\begin{longtable}{@{}p{0.14\textwidth}p{0.19\textwidth}p{0.16\textwidth}p{0.17\textwidth}p{0.14\textwidth}@{}}
\caption{Supported claim category by setting. Entries interpret frozen category transitions and are not pooled scores.}\\
\toprule
Setting & Instance-set improvement & Fidelity preservation & Safety preservation & Evidence stability \\
\midrule
\endhead
Source development & 2/3 seeds complete; seed 41 below F1 threshold & Not independent confirmation & Not complete in seed 41; empty activation increased & Progression evidence only \\
Direct source test & 3/3 seeds complete & 3/3 met the frozen union-Dice condition & 3/3 had no empty-frame increase; solver flags zero & Stable within sealed contract \\
Direct Endoscapes & 1/3 complete; stable transfer not established & 1/3 complete; seeds 17 and 41 failed the F1/fidelity conjunction & Preservation not established; seed 17 had an empty-frame increase & Seed-unstable direct transfer \\
Direct ROBUST-MIPS & 3/3 seeds complete & 3/3 met the frozen union-Dice condition & 3/3 had no empty-frame increase; solver flags zero & Complete within the ROBUST-MIPS contract \\
\bottomrule
\end{longtable}
\normalsize

Failure to establish safety preservation is not evidence that safety degradation occurred.

\section{Exploratory mechanism diagnostics}
\subsection{True cardinality and shared-cardinality crossover}
The true-cardinality report contains 12 complete domain--seed--arm records. Count MAE, solver flags, and empty false activation were zero in the reported control outputs. Descriptive external means did not show a stable positive graph-over-node relation under fixed cardinality. The graph was not consistently best across Endoscapes or ROBUST-MIPS seeds. These descriptive results do not identify a fixed-cardinality relation-specific gain.

The shared-cardinality descriptive rows gave F1 values 0.7656, 0.7540, and 0.7481 for graph-cardinality with graph, capacity, and shuffled selectors; 0.7142, 0.7042, and 0.7016 for capacity-cardinality; 0.7489, 0.7414, and 0.7374 for shuffled-cardinality; and 0.7761, 0.7785, and 0.7693 for true-cardinality. The selector spread was 0.0175 versus a cardinality-source spread of 0.0619. The pattern is consistent with a substantial cardinality contribution, but it does not prove mediation or isolate a relation-specific residual.

\subsection{Capacity, shuffled, and candidate-coverage controls}
The capacity-matched node control had 59,334 parameters versus 59,120 in the graph path. The reported anchors included a safety failure where empty activation changed from 8 to 10, so the follow-up rule did not pass. Shuffled and cross-axis controls remain descriptive because capacity and selector effects are not jointly isolated.

In the source-test candidate-missingness atlas, 135 of 1,553 positive ground-truth instances were missing from the candidate pool (0.087); the missing targets had mean area 716.3 versus 3,482.1 for covered targets. In a seed-23/fold-1/g101 high-resolution anchor, candidate ceiling rose from 0.8831 to 0.9436, but downstream F1 at replacement quota 20 fell from 0.7189 to 0.6853 and empty activations rose from 7 to 9. This separates candidate coverage from final utility.

\subsection{Calibration, presence, and A2}
Calibration did not provide a stable joint repair across the reported controls, and presence controls failed when lower empty activation was accompanied by positive loss-budget violations. A2 ended at Stage A: pair-supervision removal gave $\Delta$F1 0.0057, set-failure reduction 0.0121, and $\Delta$Dice -0.0020; cardinality-supervision removal gave $\Delta$F1 -0.0118, set-failure reduction -0.0068, and $\Delta$Dice -0.0022. Stage B was not run. None of these controls identifies a single component effect.

\section{Supplemental held-out-domain relation}
The supplemental leave-one-domain-out analysis did not support multi-domain held-out robustness. It is a different training/evaluation relation from the direct settings and is reported only to bound generality. Where empty false activation increased, the interpretation remains that safety preservation was not established, not that safety degradation was supported.

\end{document}